\documentclass{article}
\usepackage{arxiv}

\usepackage{arxiv}
\usepackage[utf8]{inputenc} 
\usepackage[T1]{fontenc}    
\usepackage{url}            
\usepackage{booktabs}       
\usepackage{amsfonts}       
\usepackage{nicefrac}       
\usepackage{microtype}      
\usepackage{graphicx}
\usepackage[linesnumbered,ruled,vlined]{algorithm2e}
\usepackage{amsmath}
\usepackage{subfig}
\usepackage{caption}
\usepackage{booktabs,siunitx}
\usepackage{wrapfig,lipsum}
\usepackage{multirow}
\usepackage[]{nohyperref} 
\usepackage{url} 
\usepackage[dvipsnames,tabularx]{xcolor}

\title{Progressive Domain Adaptation with Contrastive Learning for Object Detection in the Satellite Imagery}

\author{
 Debojyoti Biswas \\
  Computer Science \\
  Texas State University \\
  San Marcos, TX 78666\\
  \texttt{ubq3@txstate.edu} \\
  \And
  Jelena Te\v{s}i\'{c} \\
  Computer Science\\
  Texas State University\\
  San Marcos, TX 78666 \\
  \texttt{jtesic@txstate.edu} \\
}

\begin{document}
\maketitle
\begin{abstract}
State-of-the-art object detection methods applied to satellite and drone imagery largely fail to identify small and dense objects. One reason is the high variability of content in the overhead imagery due to the terrestrial region captured and the high variability of acquisition conditions. Another reason is that the number and size of objects in aerial imagery are very different than in the consumer data. In this work, we propose a small object detection pipeline that improves the feature extraction process by spatial pyramid pooling, cross-stage partial networks, heatmap-based region proposal network, and objects localization and identification through a novel image difficulty score that adapts the overall focal loss measure based on the image difficulty. Next, we propose novel contrastive learning with progressive domain adaptation to produce domain-invariant features across aerial datasets using local and global components. We show we can alleviate the degradation of object identification in previously unseen datasets. We create a first-ever domain adaptation benchmark using contrastive learning for the object detection task in highly imbalanced satellite datasets with significant domain gaps and dominant small objects from existing satellite benchmarks—the proposed method results in up to a 7.4\% increase in mAP performance measure over the best state-of-art. 
\end{abstract}


\begin{figure}[!ht]
\centering
    \includegraphics[width=.80\textwidth]{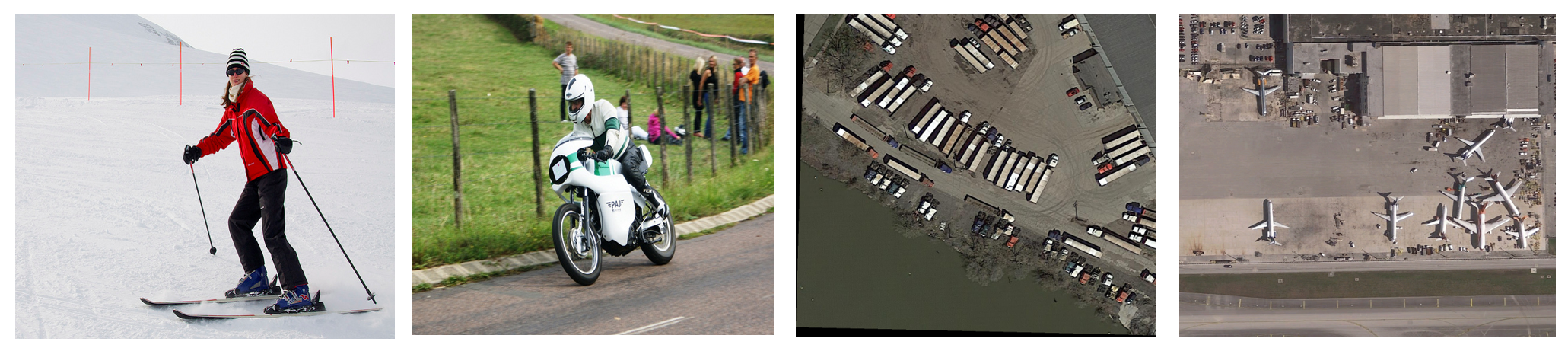}
    \caption{Visual difference between consumer images\cite{lin2014microsoft} and aerial images \cite{2020LiObject}} \label{fig:Ex_natural_aerial}
\end{figure}

\section{Introduction}
\label{sec:Intro}

\begin{figure*}[!ht]
\centering
    \includegraphics[width=.50\textwidth]{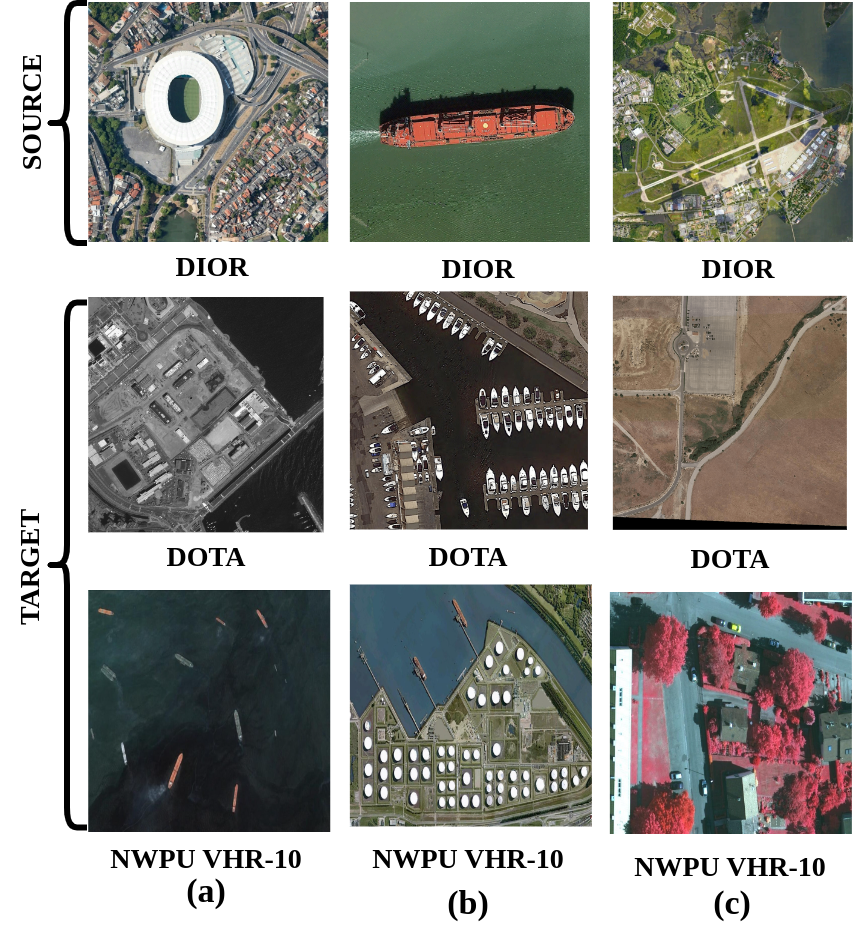}
    \caption{Illustration of variations from different aspects in our experimental datasets: (a) lighting conditions, (b) object shape and scale, and (c) variation due to geographical change} \label{fig:variations}
\end{figure*}

There is a growing need for automated object localization and identification systems for overhead imagery for traffic control, national parks, wilderness areas, natural disaster surveillance, agriculture, maritime piracy, etc. Research efforts are underway in precision agriculture  \cite{valente2020automated}, emergency rescue \cite{workman2020dynamic}, terrestrial and naval traffic monitoring \cite{9925528}, and industrial surveillance \cite{liu2021boost} to integrate accurate automated object localization and identification in overhead systems. The challenge lies in the fact that due to high ground sample distance (GSD), the aerial imagery content varies significantly within the same area of capture or drone flight. Several factors are responsible for this dramatic change, such as significant changes in light conditions (time of day, season, weather) and the type of terrestrial terrain captured in the imagery. The variation between datasets, including multiple dates, terrains, missions, object distribution and sizes, and lighting conditions, is even higher. Figure \ref{fig:variations}, (a) show the variations due to image capture time and lighting conditions, Whereas Figure \ref{fig:variations}(b) illustrate that an object can be as small as 0.01\% and as large as 70\% of an image. Figure \ref{fig:variations}(b) also shows a loosely packed nature vs. the densely packed small object characteristics in aerial images. Further, we claim that geographical variance is a critical challenge for domain adaptation tasks on satellite images in Figure \ref{fig:variations} (c); we also see that geographical variance can also exist in a singular experimental dataset due to image capture from regions of the world. In figure \ref{fig:variations}, the domain gap between the source and target datasets from different aspects is also noticeable by investigating row 1 (source dataset) with rows 2 and 3 as the target datasets.  

On the other hand, examples of consumer images vs. overhead images are illustrated in Figure~\ref{fig:Ex_natural_aerial}.  Advancements in deep neural networks and greater availability of computational resources have led to enhanced object detection techniques in consumer images \cite{long2020pp}, and the improvements include improving the Region Proposal Network, better backbone, and integration of the weight-based loss function for hard-example mining. State-of-the-art object detection and domain adaptation (DA) modeling approaches developed for consumer images do not translate to overhead imagery due to visual variation within the image, variation among the images in the collection, the relative object size w.r.t the image, the image size, and the density and number of objects in an image \cite{mittal2020deep}. Recent advances in model detection in overhead images do not address the domain shift and the challenges due to high GSD in an unsupervised setting with a single pipeline \cite{zhu2021tph}. The object identification models for UAV/drone images tend to perform better in low-altitude datasets with relatively fewer small and dense objects compared to our experimental datasets \cite{srivastava2021survey}.

 \begin{figure*}[!ht]
\centering
    \includegraphics[width=.95\textwidth]{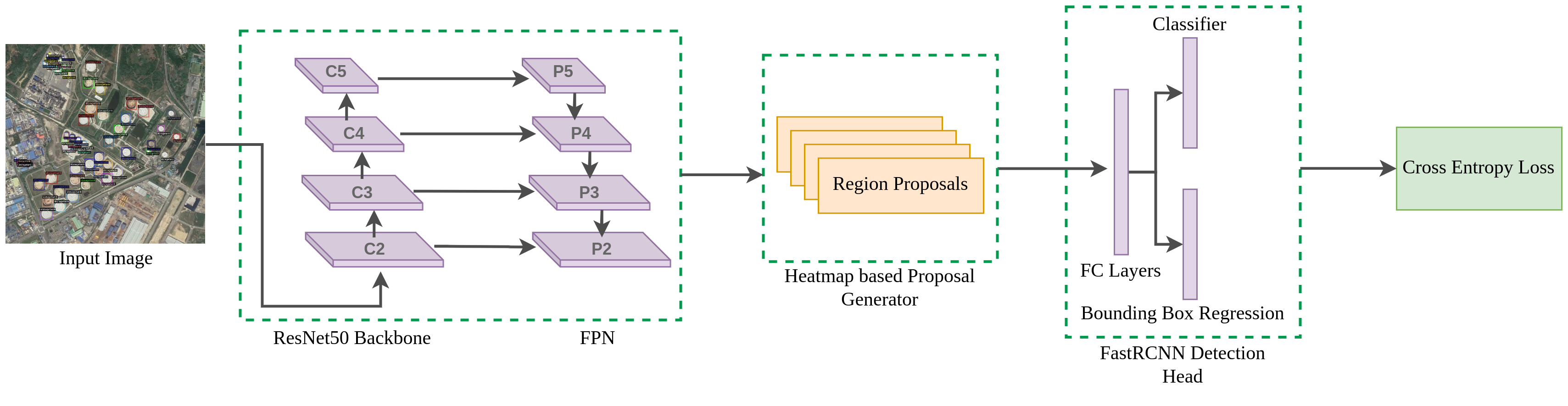}
    \caption{\emph{CenterNet2}: Heatmap based multi-stage small object detection model used as a baseline \cite{zhou2021probabilistic}} \label{fig:Base}
\end{figure*}

Deep neural networks require large and diverse amounts of annotated training data to guarantee reliable object localization in unseen datasets. Collecting and annotating aerial datasets has proven difficult and complex due to many small and dense objects per image \cite{lam2018xview}.  Only three datasets with rich class distribution are being used to benchmark the results to date: DOTA2.0 \cite{2021Ding}, and DIOR \cite{2020LiObject} are two satellite image collections, and VisDrone is the overhead drone collection \cite{zhu2018vision}. Domain adaptation successfully shares acquired knowledge (regarding annotations and learned models) in the source domain with the target domain. Domain adaptation has risen as one of the approaches to speed up the pseudo-labeling of objects in the target domain using source labels. Domain adaptation for object recognition in consumer image datasets successfully addresses weather, lighting conditions, geological variance, variation in image quality, and cross-camera adaptation by aligning the global feature distribution of data from the origin and target domains \cite{deng2022hierarchical}. Recent State-of-the-art (SOTA) work of unsupervised domain adaptation for aerial imagery uses the reconstructed feature alignment method instead of adversarial-based feature alignment to avoid background noise alignment \cite{zhu2022rfa}. Nevertheless, limited progress in the domain adaptation task has been focused on satellite imagery. Therefore, there is a pressing need to delve into new ideas and techniques to achieve more improved and favorable results. 

In this work, we intend to explore contrastive learning on local and global image features to perform feature alignment on task-specific layers. Contrastive learning is a technique that evaluates pair-to-pair relationships by measuring the similarities between different pairs, such as query-positive or query-negative. It groups similar features closely and dissimilar features at a distance in the feature space. We use a random image feature as the query sample and its augmented version as the positive sample. Negative samples are the image features in a mini-batch not part of the query and positive samples. This paper introduces the first domain adaptation benchmark for large-scale satellite image datasets. To reduce the global gap between the source and target domains, we create two intermediate domains using the CycleGAN modeling \cite{CycleGAN2017}. Then we extract the local and global feature extraction from feature pyramid network (FPN) layers using the adapted domain adaptation approach. Next, we introduce difficulty-weighted focal loss (DWFL), which uses the number of foreground proposals and amount of neuron activation and assigns a difficulty score for a particular image. Finally, we introduce the noise-contrastive estimation (InfoNCE) \cite{oord2018representation} loss to produce domain-invariant features. The methodology is outlined in Section~\ref{sec:Method}, and the novelties of the proposed methods are: 
\begin{enumerate}
\vspace*{-0.3em}
\item Using the local-global feature alignment from the source and target datasets using contrastive learning-based domain adaptation.
\item Integration of a novel difficulty estimation method in the domain adaptation pipeline.
\item \emph{Multiple} numbers of negative samples for debiased contrastive learning and object detection tasks. 
\item Progressive domain adaptation by creating an intermediate domain and minimizing the domain gap between source and target datasets. 
\vspace*{-0.3em}
\end{enumerate}

The rest of this article is organized as follows. Section~\ref{sec:Related} summarizes related work, and Section~\ref{sec:SOD} introduces the proposed methodology for the difficulty-based small object detection and training pipeline. Next, Section \ref{sec:Method} describes the contrastive learning approach and the different DA modules in the pipeline. In Section~\ref{sec:Exp} we evaluate the proposed framework using the latest cross-domains detection benchmarks over three different high-altitude (DIOR, NWPU VHR-10, and DOTA2.0) remote sensing datasets and finally summarize the findings in Section~\ref{sec:Con}.

\section{Related Work}
\label{sec:Related}

\textbf{Object Detection} The latest object detection techniques are classified into single-stage or multistage detectors. As the name suggests, single-stage detectors aim to predict object bounding boxes and class labels directly from a single network pass \cite{redmon2016you, zhu2021tph, bochkovskiy2020yolov4}. Single-stage detectors do not have separate neural network modules to generate object proposals and rely on anchor boxes of varying scales and aspect ratios. Single-stage object detection architectures may struggle with accurately localizing small or densely packed objects due to the limited receptive field of the network \cite{2021Ding, liu2021survey}. Multistage detectors are often more accurate and computationally expensive than single-stage detectors due to their use of region proposal networks (RPN) and the non-maximum suppression technique (NMS) to refine the regions of interest in images \cite{jiang2021semantic, gupta2019lvis}. NMS filters out positive instances by rejecting overlapped proposal regions in the image with the help of IOU \cite{ren2015faster, he2017mask} threshold. CenterNet2 \cite{zhou2021probabilistic} is a heatmap-based two-stage approach with balanced positive/negative samples per batch. It uses a Gaussian filter to create a heat-map peak at the object's center to define the proposal regions \cite{zhou2021probabilistic}. The anchor to the object is the region's center based on location. Thus, the one anchor per object eliminates the need for the non-maximum suppression filtering of the overlapping proposals without affecting the quality of the proposal. Due to its superior characteristics in finding densely packed and small objects, we chose CenterNet2 as our baseline architecture.

\textbf{Small and dense Object Detection} As the object's size decreases, the chances of losing local information in deep layers increase significantly. The outcome of the small object detection depends on how well the backbone network \cite{he2016deep, redmon2018yolov3} captures the region features from the input image. Next, different scale features from various stages of the backbone have been successfully used for other scale predictions. The feature pyramid network (FPN) layer also helps to strengthen standard spatially rich features by combining semantically rich features by combining low-level and high-level features with the fuse connection and up-sample method. An improved FPN module uses a similarity-based fusion method capable of extracting information for various sizes of instances \cite{zhu2022improved}. Authors argue \cite{liu2022global}, pixel-level appearance features do not contain enough information to localize small objects in an image, the global context aggregation module and the feature refinement module to build Global Context-Weaving Network are required for optimal performance in small object detections \cite{wu2022gcwnet}. Hence, context-based feature extraction is more robust for complex object and scene detection and performs better in benchmark datasets \cite{zhang2022attention}. Yang et al. \cite{yang2022querydet} propose \emph{querydet}, which first predicts the coarse locations of small objects on low-resolution features and then computes the accurate detection results using high-resolution features sparsely guided by those crude positions. This work does not rely much on low-level feature queries because it is hard to distinguish small objects with very low resolution on feature maps. The above discussion verifies that small object detection requires global/high-level features to classify objects accurately, which motivates us to use features with high receptive field from later layers in conjunction with the Multi-layered perceptions (MLP) module to calculate the difficulty score for an image (see details in section \ref{sec:SOD}).

\begin{figure*}[!ht]
   \centering
    \includegraphics[width=.95\textwidth]{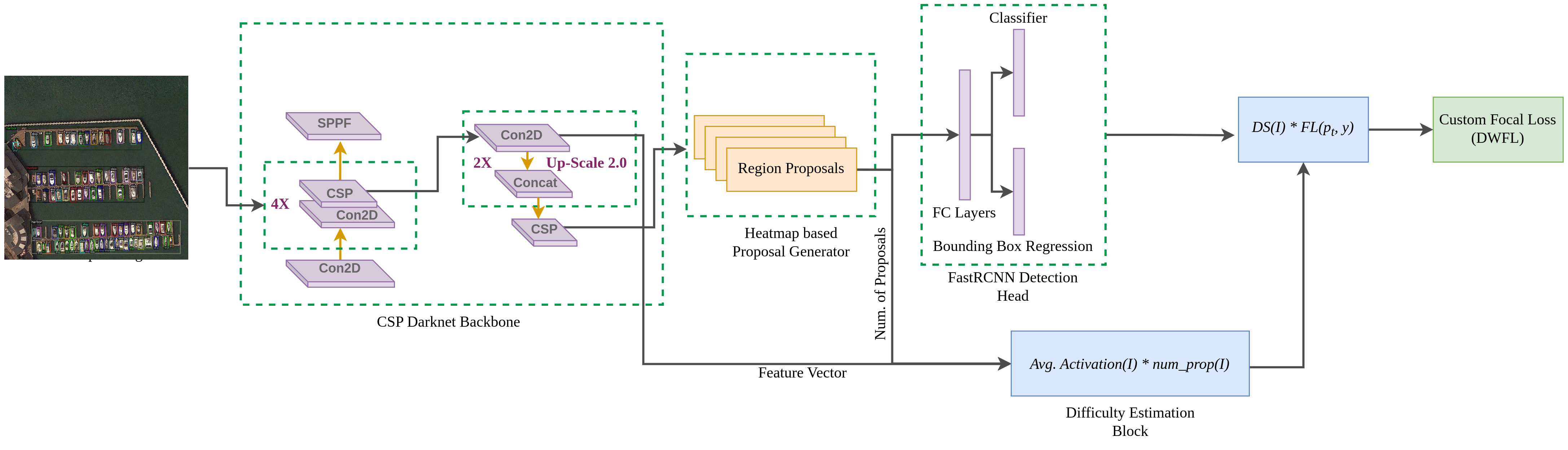}
    \caption{\emph{SOD} model: {S}mall {O}bject {D}etection model with updated backbone, new loss function, and the difficulty estimation block.}
    \label{fig:SOD}
\end{figure*}

\textbf{Domain Adaptation for Object Detection}
Domain Adaptation techniques are used to handle the problem of domain change between source and target data sets. In the last few years, the unsupervised Generative Adversarial Network (GAN) has been critical in solving the domain shift problem. The GAN-based approach expands object detection in consumer images to other domains. The GAN-based domain adaptation model uses Gradient Reversal Layer (GRL) to learn domain invariant features. Hsu et al.\cite{hsu2020progressive} use adversarial learning to align the distribution of characteristics between domains and perform progressive domain adaptation to address the problem of significant domain gaps. Saito et al. show domain adaptation should not be rigorous and uniform at every point in feature space, and careful inspection is required to align features that are very close to the inter-class boundary to achieve optimal results \cite{saito2019strong}.

On the other hand, the maximum mean and central moment discrepancy approaches successfully produced domain-invariant features through the alignment of feature space. Long et al. proposed DAN \cite{long2015learning}, which matches the mean embedding of different domain distributions from different task-specific layers in CNN, and Zeillinger et al. \cite{zellinger2017central} proposed to use means of order-wise moment differences to match the higher-order central moments of probability distributions. State-of-the-art guides knowledge transfer between domains while maintaining consistency of the relevant semantics before and after adaptation \cite{hoffman2018cycada}. Class-level distribution alignment across the source and target domains was achieved using the Easy-to-Hard Transfer Strategy and a Prototype Feature Alignment Network \cite{chen2019progressive}. However, some adversarial adaptation methods aim to mitigate the need for uniform alignment among all samples; their effectiveness remains constrained by the challenge of achieving accurate unsupervised adaptation.

\textbf{Contrastive Learning for Domain Adaptation}
Apart from the traditional GAN networks, contrastive learning \cite{hadsell2006dimensionality} in domain adaptation has gained much attention due to its straightforward work process to produce similar features across domains. However, it maintains discriminating characteristics in task-specific features to represent different classes in a singular domain. It uses a similarity function such as cosine similarity or Euclidean distance to measure the similarity between two vectors. Then it represents the image features in the domain invariant feature space. 

There are several versions of InfoNCE \cite{oord2018representation} loss for contrastive learning. Some of them focus on the number of negative examples we need for designing optimal contrastive learning. Introducing a large number of negative examples improves performance, and Wu et al. \cite{wu2021rethinking} argued that real-world datasets often introduce noise, and incorporating too many noisy negative examples yield sub-optimal results. Chuang et al. \cite{chuang2020debiased} propose reducing bias in contrastive learning by carefully selecting negative examples. The selection of \emph{False Negative} examples can disturb the learning process and harm the overall performance. Contrastive learning is successful not only in single-source domain adaptation but also in multi-source domain adaptation \cite{kang2020contrastive}. Kang et al. \cite{kang2020contrastive} consider class information and label the target dataset using the K-means clustering method. Kalantidis et al. \cite{kalantidis2020hard} use \emph{hard negative mixing} strategy to amplify the effect of negative samples with very minimum overhead computation. Motivated by these works, we choose to use contrastive learning for the domain adaptation task. We present contrastive learning that is less susceptible to False Negatives from highly imbalanced datasets by carefully selecting negative examples.

\section{Small Object Detection}
\label{sec:SOD}

The satellite image has a maximum of $400$ million pixels and objects are frequently smaller than 100 pixels. A typical patch of the image is $1024 \times 1024$, which equals $1.05$ million pixels. If an object is $10 \times 10$ or $100$ pixels, the object's size is $<0.0001$ of the area of the image. The success of object detection is contingent on the reliability of the pixel- and object-feature extraction, as well as the RPN-based proposal network within the DNN architecture. An increase in the number of small, densely packed objects raises the possibility of losing pixel-level information during feature extraction. The RPN-based proposal network can miss small objects early in the processing, leading to difficulty in detection at later stages \cite{tong2020recent}. Furthermore, dense object arrangements result in extra noise from surrounding information during input and more post-processing operations.

\subsection{Baseline Model for SOD} 
\label{ssec:base}

Figure \ref{fig:Base} shows the \emph{Base} model architecture, a pipeline adopted from the CenterNet2 model \cite{zhou2021probabilistic} with 3 components: Backbone, RPN, and Detection Head. To enhance performance on overhead datasets, the image size, output channels, and IOU were optimized in the FastRCNN Detection Head. The \textbf{Backbone} employs ResNet50 as a feature extractor and FPN for multi-scale predictions, combining features from prior layers via residual connections to prevent vanishing gradients problems. The Base model, with ResNet50, achieved leading results on COCO \cite{lin2014microsoft} and LVIS \cite{gupta2019lvis} datasets. The FPN layer extracts features at different scales from different Backbone layers, shown in the FPN block in Figure \ref{fig:Base}. The Resnet(C3), Resnet(C4), and Resnet(C5) blocks represent strides of 8, 16, and 32 in the network. P3, P4, and P5 show three scale prediction FPN outputs, which are fed into the RPN block described in the following paragraph for scale predictions.

 \textbf{Region Proposal Network (RPN)} in the \emph{Base} model \cite{zhou2021probabilistic} creates region suggestions through heatmaps by applying Gaussian kernels on features from the FPN at different scales. The heatmaps are produced by comparing the max-pool input and the Gaussian kernel output element-wise. Max-pooling highlights each pixel in the feature except for local maxima, which have a value of 1. Each peak in the heatmap represents the center of an object, as shown in Figure \ref{fig:detections} (a) and (c). The features at each key point are used to determine the size of objects, leading to accurate bounding boxes even when objects are close or overlap. To improve performance on overhead imagery, the baseline model requires advanced \emph{image augmentation} techniques, enhanced \emph{feature extraction} methods, and an increased \emph{proposals per image} and \emph{detection per image}. \textbf{Detection Head} is based on the Faster-RCNN detector \cite{ren2015faster}. It takes filtered region proposals from the RPN as input and processes them as follows: 1) Convert proposals into $7 \times 7$ grids with the same number of channels via region-of-interest (ROI) pooling, 2) Flatten the pooling output and feed it into FCN for final detection output, $(N,C)$ for class predictors for $C$ classes, and $(N,4)$ for $N$ region proposals bounding boxes.

\begin{figure*}[!ht]
   \centering
    \includegraphics[width=.95\textwidth]{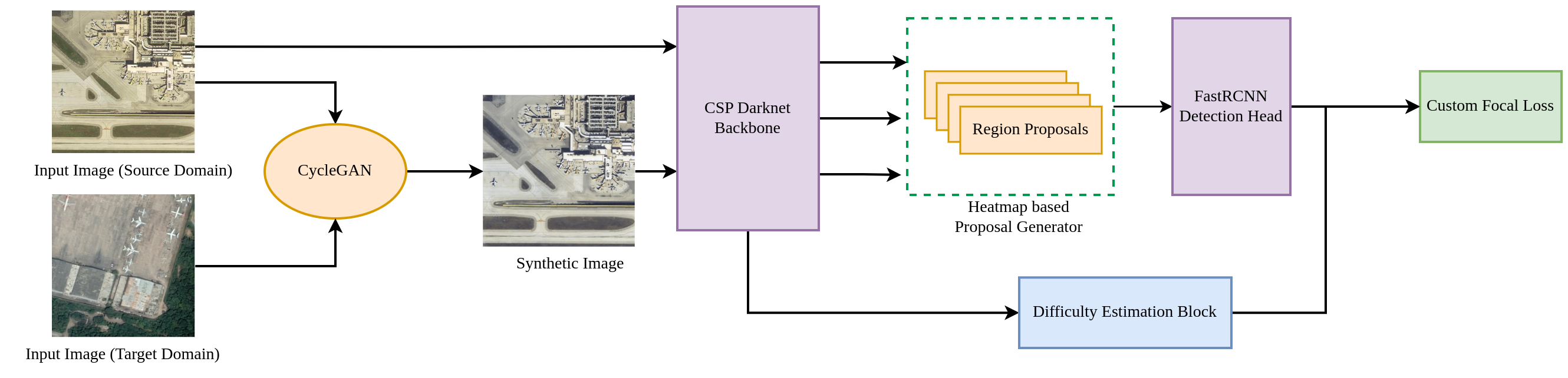}
    \caption{\emph{HeatDA}: Heat Domain Adaptation model with transfer learning and CycleGAN translated image domain adaptation.}
    \label{fig:heatDA}
\end{figure*}

\subsection{Small Object Detection pipeline} 
\label{ssec:Base}

An extended version of our \emph{SOD} model for small object detection \cite{biswas2022small} improves upon the Base model by optimizing the pipeline for detecting small objects. We replaced the backbone with CSP Darknet \cite{bochkovskiy2020yolov4} and added the Difficulty Estimation block. We also switched to a modified focal loss instead of cross-entropy loss (Figure~\ref{fig:SOD}). The RPN module is effectiveness depends on the backbone's performance. If the backbone fails to extract meaningful features for the small object in the image, the RPN module will likely fail to include the small object in the region proposals. Our findings found that 75\% of RPN proposals in the Base model were trivial and repetitive (backgrounds, partial objects) as explained in \cite{biswas2022small}. To enhance the model, we used the CSP Darknet backbone for preserving better semantic information in deeper CNN layers \cite{bochkovskiy2020yolov4, wang2020cspnet}. The system introduces a partial cross stage for low/high-res aggregation and replaces max-pooling with spatial pyramid-pooling for finer feature extraction. 

\begin{figure*}[!ht]
    \centering
    \includegraphics[width=.9\textwidth]{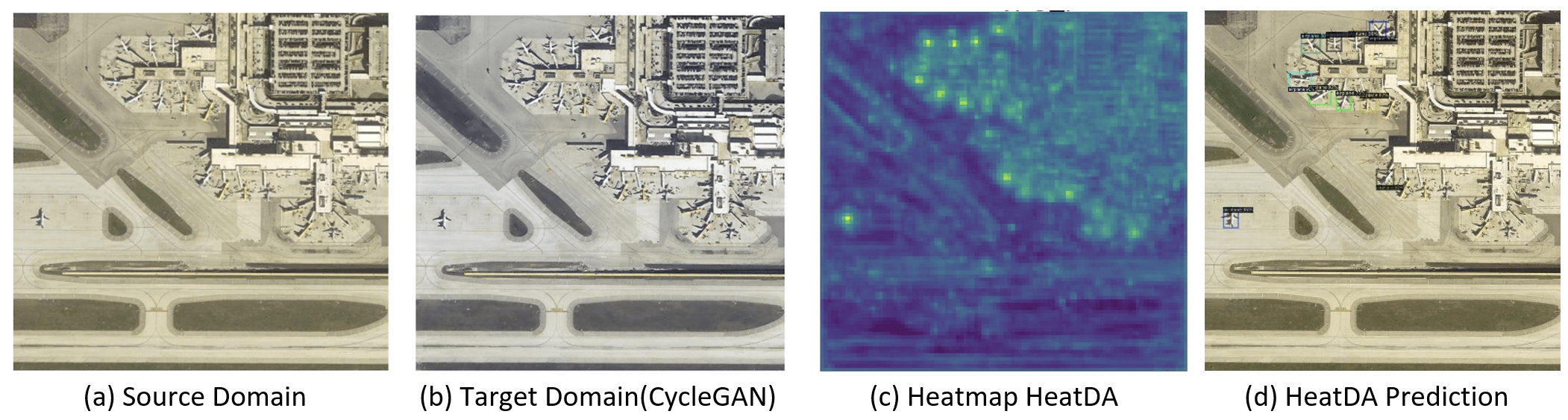}
    \caption{Domain translation from Source(S) to Source as Target(SaT) using CycleGAN \cite{CycleGAN2017} and Object Detection from HeatDA model.} \label{fig:HeatDA_detect}
\end{figure*}

The proposed small object detection model in Figure~\ref{fig:SOD} concatenates several low-level features with high-level features channel-wise, thus propagating the semantic information from the lower to higher levels. The region proposal network considers the multi-resolution features from different CSP-Bottleneck layers for proposal generation, as illustrated in Figure~\ref{fig:SOD}. \textbf{Difficulty Estimator (DE)} module numerically captures the complexity of an image feature based on the number of foreground object proposals and the amount of neuron activation information in the network for an image. It calculates the overall difficulty of each image by taking feature input from different stages of the feature network. The difficulty score (DS) for an FPN feature level with a resolution of $C \times W \times H$ for the image $I$ is calculated in Eq.~\ref{eq:DS}. 

\begin{equation}
DS(I)= \frac{1}{L} \sum_{l=1}^{L} (\frac{1}{C*W*H} \sum_{c=1}^{C} \sum_{w=1}^{W} \sum_{h=1}^{H} f_{c,w,h}(I)) 
* \text{ } num\_prop(I)
\label{eq:DS}
\end{equation}

The feature output channels, width, and height at FPN level (l) are represented by \emph{C, W, H} respectively. In contrast, L represents the number of FPN levels used to calculate difficulty. The difficulty score (DS) at the FPN level is calculated by dividing the sigmoid linear unit (SiLU) activation values $f_{c,w,h}(I)$ at all pixels in an image (I) by the total dimension of C, W, H. Using this block we calculate the number of total neurons fired for a single image in the forward pass. DS is derived from three FPN levels, averaged, and multiplied by the number of proposals (See Eq. \ref{eq:DS}). The final DS value for an image (I) is obtained by normalizing the DS value between $[0.5, 1.4]$. The complexity increase for the DS block is minimal, and its computational time, expressed in Big Oh \emph{(O)} notation, is \emph{O(r)}, where r is the number of iterations during training.

\begin{equation}
    FL(p_t,y)= \alpha_t * (1-p_t)^\gamma * CE(p, y) ,\\
  DWFL(x, p_t, y)= DS(I) * FL(p_t, y),
  \label{eq:focal}  
\end{equation}

\textbf{Difficulty weighted Focal Loss} is calculated from the difficulty scores for each image, and we propose replacing the loss of cross-entropy with the loss of custom focalization, as illustrated in Figure~\ref{fig:SOD}. The difficulty scores are calculated using Eq. \ref{eq:DS} for each image by a difficulty estimator block as a weight factor to focus more on complex images with many small objects and a high variation in pixel-level features. The basic form of the focal loss function is outlined in Eq. \ref{eq:focal}.

\begin{equation}
\forall c \in C, \alpha'_c = -1 * log \left( \frac{|C_c|}{|C_1 \cup C_2 \cup ...|} \right) 
\Rightarrow \alpha_c = \beta*\frac{\alpha'_c-min(\alpha_c)}{max(\alpha_c)-min(\alpha_c)}
\label{eq:alpha}
\end{equation}

The $p_t$ is the probability distribution of the target \emph{t}, and $y$ is the ground truth of the object being a specific class, $\gamma$ is the modulating factor, $\alpha_t$ is used as a weighting factor, and CE represents the cross-entropy function. We propose a new measure, the Difficulty Weighted Focal Loss (DWFL) in Eq.~\ref{eq:focal} as a product of difficulty score, $DS(I)$ in Eq.~\ref{eq:DS}, and focal loss for the image, $FL(p_t,y)$ in Eq.~\ref{eq:focal}. The value $\alpha$ is used in the $FL(p_t,y)$ calculation to control the class imbalance problem in our source and target data sets. The $\alpha_c$ is calculated as in Eq.~\ref{eq:alpha} for each class, where the modulating factor $\alpha'_c$ depends on the frequency $|C_c|$ of a particular class in the data set and $|C_1 \cup C_2 \cup C_3 ...|$ is the total number of all instances of all classes in the data set. The normalized $\alpha_c$ values from Eq. \ref{eq:alpha} are used across different classes $c$, $c \in C$ to mitigate the imbalance of object class labels. In the experiment section~\ref{sec:Exp} we confirm that the proposed normalization of $\alpha_c$ in Eq.~\ref{eq:alpha} is more effective and gives a stable loss calculation for a highly unbalanced class count in the data set. 

\section{Domain Adaptation Methods}
\label{sec:Method}

Different overhead image datasets are usually taken at different geographical locations, and different types of satellites were used to capture images with different orientations under various weather and lighting conditions. There are a handful of annotated overhead image collections \cite{wang2022remote}, and they all have different object class annotations, both in frequency and assigned object labels. This contributes to a large domain gap between our source and target data sets. Object detection performance on target degrades drastically when the domain gap is very w.r.t source dataset. Domain adaptation (DA) methods are key to solving this problem. Using domain adaptation methods, we can perform better in unseen datasets not introduced during the training phase. The self-supervised or unsupervised domain adaptation aims to produce invariant features for a particular class across domains. In the experiment section~\ref{sec:Exp}, we confirm that the domain adaptation of the source in the training process improves the object detection performance. 

\subsection{\emph{HeatDA} model: Heat Domain Adaptation Model} 
\label{ssec:heatDDA}

Here, we propose a pre-processing step for the pipeline outlined in Figure~\ref{fig:SOD} and map the source domain into the target domain first, as illustrated in Figure~\ref{fig:heatDA}, as we have found that closing the source and target gap using progressive domain adaptation leads to better object detection in the previously unseen overhead imagery. Using source and target image examples, we train the CycleGAN domain discriminator \cite{CycleGAN2017, isola2017image}. The resulting domain discriminator model translates the source image to the target domain, as illustrated in Figure~\ref{fig:HeatDA_detect}. This additional CycleGAN domain discriminator model illustrated in Figure~\ref{fig:heatDA} allows us to align pixel-level features between two domains and use the source image and the translated source image to train the SOD network, as illustrated in Figure~\ref{fig:heatDA}. The conversion of the source to the target domain allows us to incorporate target-like domain characteristics without relying on the object-level annotations that might or might not be present. Training the \emph{HeatDA} using target-like images helps to align pixel-level semantic information for the source and target domain, thus improving the detection performance (see Figure \ref{fig:HeatDA_detect}) in the target dataset.

\begin{figure}[!ht]
\centering
\includegraphics[width=0.48\textwidth]{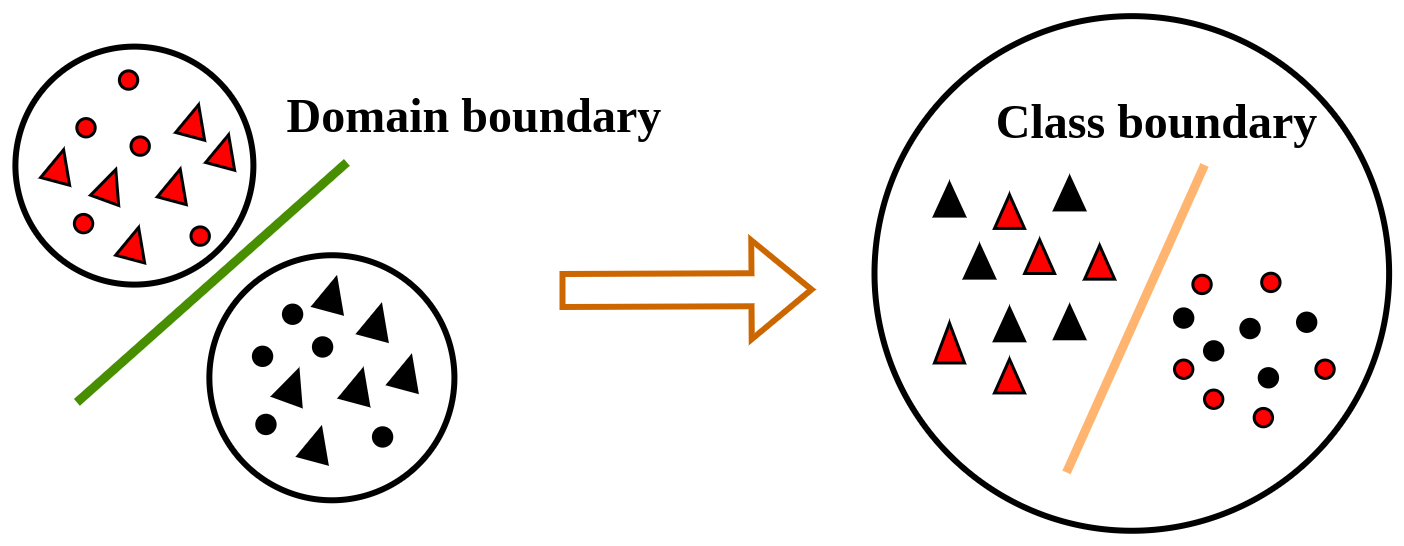}
    \caption{Domain Adaptation with contrastive learning. Here, different colors indicate different domain distributions, and the different shapes represent different classes in a domain. The Green and Orange line represents the domain and class boundary, respectively.}\label{fig:constrastive}
     \vspace{-1.0em}
\end{figure}

\begin{figure*}[!ht]
\centering
\includegraphics[width=0.9\textwidth]{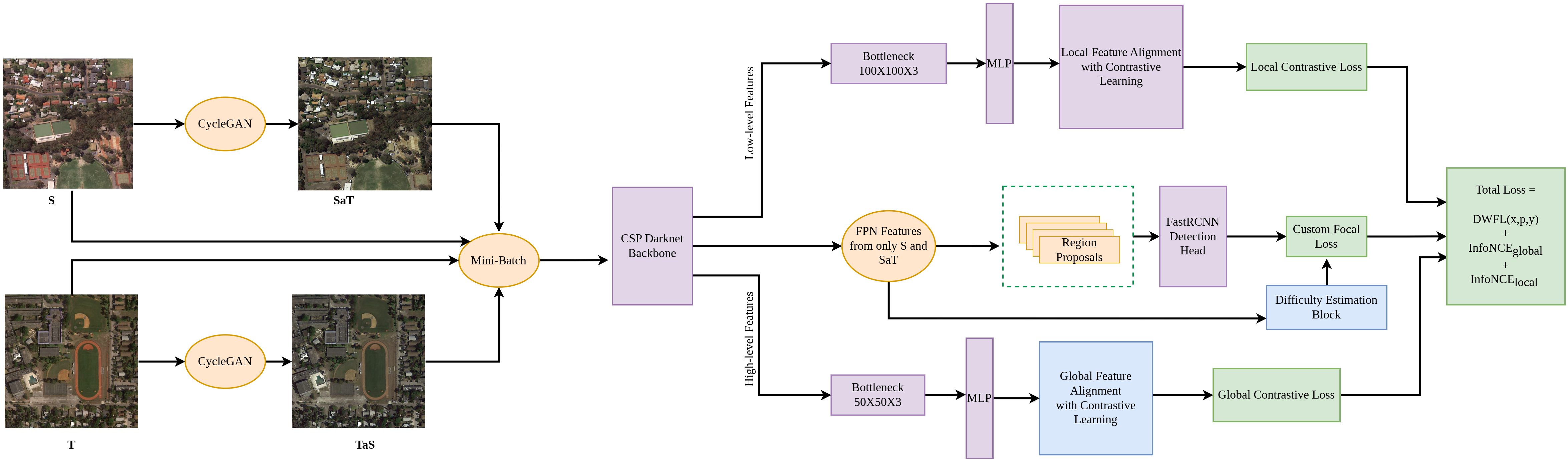}
    \caption{Local and Global Domain Adaptation(LGDA) model with contrastive learning.} \label{fig:LGDA}
\end{figure*}

\subsection{\emph{LGDA} Model: Local Global Domain Adaptation Model} 
\label{ssec:UDACons}

Contrastive learning \cite{hadsell2006dimensionality, chen2020simple} is a simple process of measuring pair-to-pair relationships based on the similarities between different pairs, such as query-positive or query-negative. Figure~\ref{fig:constrastive} illustrates the functional strategy of contrastive loss. Feature representation of the source and target objects differs in the feature space, and there is a huge gap due to lightning, geographic, weather, and acquisition differences, and the difference is illustrated by a green line in Figure~\ref{fig:constrastive}. Contrastive learning brings similar points to close together and pushes dissimilar points separate from each other by calculating similarities between pairs \cite{kang2020contrastive, deng2022hierarchical}. A pair of feature vectors with high similarity are placed close together, and vector pairs with low similarity are placed distantly in feature space. In the ideal case, the contrastive domain adaptation maps the feature space of the source dataset to the target dataset so that the features representing objects in the same class in the source and the target domain dataset are closer together. In this light, we propose to enforce contrastive learning on local features to minimize the domain gap w.r.t the local characteristics in the image i.e. color and texture captured by deep features of the pixel and its nearest surroundings. Here, we produce domain-invariant object features for source and target domains by increasing similarities between \emph{Query-Positive} pair and decreasing similarities between \emph{Query-Negative} pairs. We introduce the Informative Noise Contrastive Estimation (InfoNCE) loss measure of finding similarities and dissimilarities between features in Equation~\ref{eq:consloss}. Similarity of two features $u$ and $v$ is captured by a cosine score  $sim(u,v) =  u^T /(||u|| * ||v||)$. The \emph{Query} is from the source, the \emph{Positive} example is from the translated source domain denoted as SaT in Figure \ref{fig:LGDA}, and negative examples are with different classes than the \emph{Query} example and from the target domain. Contrastive learning increases the similarity between \emph{Query} and the \emph{Positive} sample and increases the dissimilarity between \emph{Query} and $N$ negative samples $Negative_n, n\in[1,N]$ s outlined in Eq.~\ref{eq:consloss}. \begin{equation}
\normalsize
InfoNCE=  -log \frac{exp(sim(Query, Pos)/\tau)}{\sum_{n=1}^{N}exp(sim(Query, Neg_n)/\tau)}
\label{eq:consloss}
\end{equation} The Informative Noise Contrastive Estimation (InfoNCE) loss measure is low when the similarity between the \emph{Query} and the \emph{Positive} example is high and when then the similarity between Query to all \emph{Negative} examples is low. Using this loss, we are learning domain invariant features.  N is the mini-batch size during the training phase, and $\tau$ is the temperature that controls the strength of penalties in hard negative samples. Our implementation ensures that we find a similar example as \emph{Positive} case and dissimilar examples as \emph{Negative} cases. Figure \ref{fig:LGDA} illustrates the proposed Local Global Domain Adaptation \emph{LGDA} model and incorporates the proposed approach in the small object detection framework. This architecture focuses on performing domain adaptation on a highly class-label imbalance dataset where labeled objects are small compared to the image size. We added two modules to two new modules: Local Feature Alignment and Global Feature Alignment for contrastive learning. Our proposed model takes input from four different distributions; among them, two are the source and target domains, and the other two are new intermediate domains, Source as Target (SaT) and Target as Source(TaS), from the source and target datasets, respectively, generated from the CycleGAN \cite{CycleGAN2017} network to reduce the gap between the source (S) and target (T) domains. 

As shown in Figure \ref{fig:LGDA}, mini-batch inputs are passed into \emph{CSP DarkNet} backbone. Next, features extracted from the backbone are fed into Local Feature Alignment and Global Feature Alignment modules for calculating \emph{Local Contrastive Loss} and \emph{Global Contrastive Loss}, respectively. However, RPN produces region proposals for object detection tasks from only S and SaT domain features passed from the backbone because we have ground truth for these two domains. The later part of the architecture is a traditional RCNN-style object detector with classification and regression modules.

 \begin{figure*}[!ht]
    \centering
    \includegraphics[width=.95\textwidth]{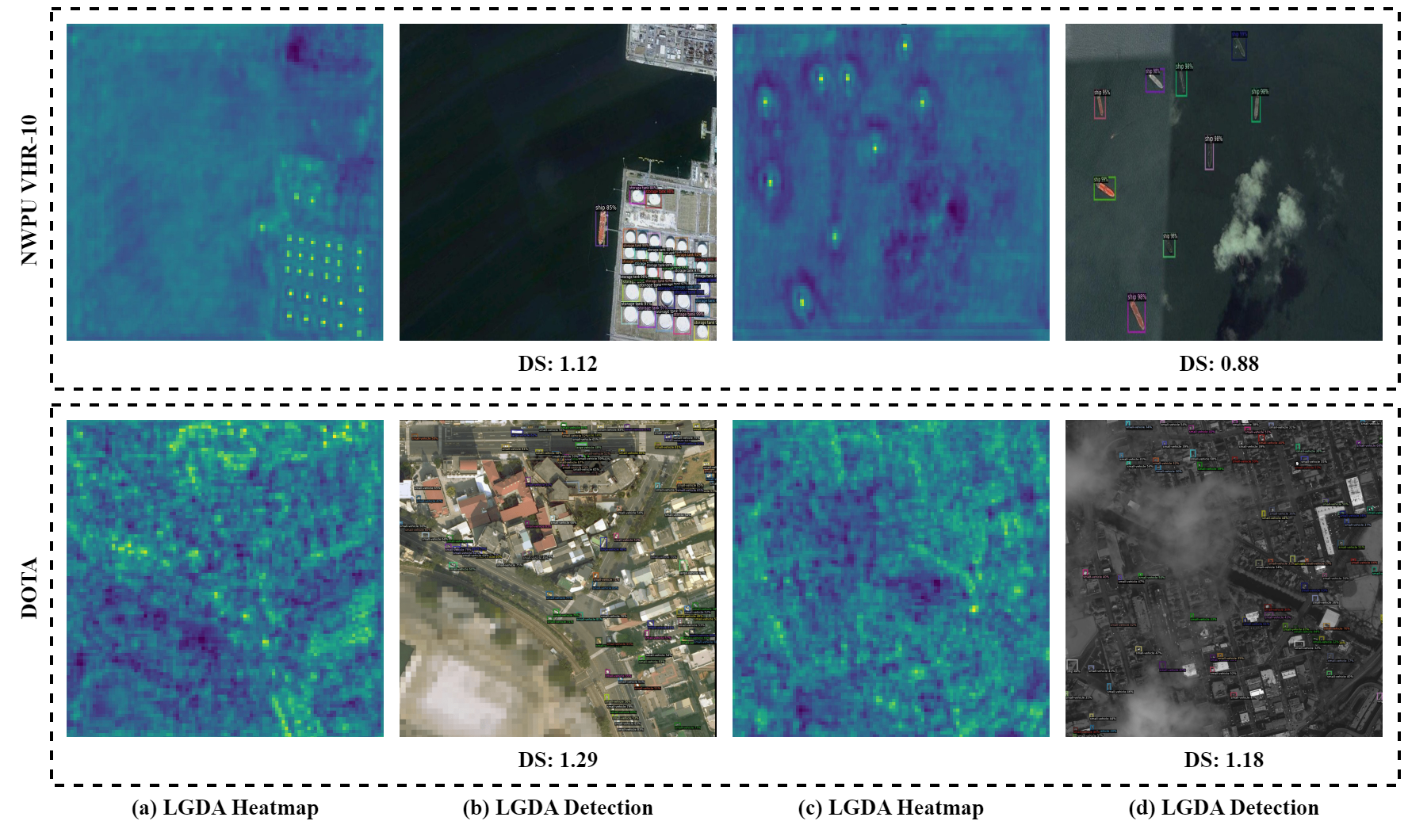}
    \caption{Proposals heatmaps and detection results from the \emph{LGDA} model for DOTA2.0 and NWPU VHR-10 target datasets. Here DS = Difficulty Score for a particular image.} \label{fig:detections}
\end{figure*}

\textbf{Local Domain Adaptation} focuses on the local features in an image and assumes there is no ground truth object labeling for the target dataset, only for the source dataset. Local features capture low-level descriptions of a pixel and its neighbors in an image. The images from a mini-batch pass through the backbone and local features are saved from the earlier layers of the backbone. The saved local features are in the shape of $256 \times 100 \times 100$, where dimensions are in the form of C, W, H, respectively. To reduce the necessity of computational power and GPU memory and improve similarity computation performance, we pass the features into the bottleneck module as shown in Figure \ref{fig:LGDA} and downgrade the shape to $3 \times 100 \times 100$. In Figure \ref{fig:LGDA}, S and T are images drawn from the source and target datasets, respectively, where SaT and TaS are the corresponding images of S and T in translated form, produced by the CycleGAN network. Let us denote the local feature vectors from \emph{S, SaT, T and TaS} as $L^S_k$, $L^{SaT}_k$, $L^T_k$, and $L^{TaS}_k$, respectively. Where $k$ is the index of the mini-batch. For the adaptation of the S and SaT domain, we select a local feature $L^S_i \in L^S$ as a query and the corresponding feature from $L^{SaT}_i \in L^{SaT}$ as the positive case. Negative cases are all other local characteristics $L^{SaT}_j \in L^{SaT}$, where $j \neq i$. The local domain loss between \emph{S, SaT} and \emph{T and TaS} are calculated in Eq. \ref{eq:StoTlocal} and \ref{eq:TtoSlocal}.

\begin{equation}
\normalsize
InfoNCE^{S, SaT}_{local}=  -log \frac{exp(sim(L^S_i, L^{SaT}_i)/\tau)}{\sum_{j=1}^{N}exp(sim(L^S_i, L^{SaT}_j)/\tau)}
-log \frac{exp(sim( L^{SaT}_i, L^S_i)/\tau)}{\sum_{j=1}^{N}exp(sim(L^{SaT}_i, L^S_j)/\tau)}, j \neq i
\label{eq:StoTlocal}
\end{equation}

\begin{equation}
\normalsize
InfoNCE^{T, TaS}_{local} =  -log \frac{exp(sim(L^T_i, L^{TaS}_i)/\tau)}{\sum_{j=1}^{N}exp(sim(L^T_i, L^{TaS}_j)/\tau)} \\
-log \frac{exp(sim(L^{TaS}_i, L^T_i)/\tau)}{\sum_{j=1}^{N}exp(sim(L^{TaS}_i,L^T_j)/\tau)}, j \neq i
\label{eq:TtoSlocal}
\end{equation}

After accumulating loss for all query images in a minibatch, the total bidirectional local domain adaptation loss can be formulated as below in Eq. \ref{eq:localLoss}. \begin{equation}
\normalsize
InfoNCE_{local}= InfoNCE^{S, SaT}_{local} + InfoNCE^{T, TaS}_{local}
\label{eq:localLoss}
\end{equation}

\textbf{Global Domain Adaptation} relies on the global alignment of features between the source and the target dataset. The global features represent a more abstract formation of objects in the image and are saved from the last layer of the backbone. The shape of the global features is $256 \times 25 \times 25$, where the dimensions are C, W, and H, respectively. Again, the same as for local features, we use a bottleneck module to reduce the size of global features to $3 \times 25 \times 25$. Global features are high-level features in the DNN pipeline. Global domain adaptation and feature alignment are also performed at the mini-batch level to restrict computational and GPU memory expense.

The global feature vectors of training mini-batch input: \emph{S, SaT, T, and TaS} are indexed as $G^S_k$, $G^{SaT}_k$, $G^T_k$ and $G^{TaS}_k$, where $k$ is the index of the mini-batch. The global contrastive loss for S and SaT is calculated by selecting a query sample $G^S_i \in G^S$ and a positive case from the corresponding image feature $L^{SaT}_i \in L^{SaT}$ and vice versa. We take the negative cases as all the other global features $G^{SaT}_j \in G^{SaT}$, where $j \neq i$, and the adaptation formula for S and SaT domains in global feature space is outlined in Eq.~\ref{eq:StoTglobal}.  
\begin{equation}
\normalsize
InfoNCE^{S, SaT}_{global}=  -log \frac{exp(sim(G^S_i, G^{SaT}_i)/\tau)}{\sum_{j=1}^{N}exp(sim(G^S_i, G^{SaT}_j)/\tau)} \\
-log \frac{exp(sim( G^{SaT}_i, G^S_i)/\tau)}{\sum_{j=1}^{N}exp(sim(G^{SaT}_i, G^S_j)/\tau)}, j \neq i
\label{eq:StoTglobal}  
\end{equation}
The adaptation of the formula for the T and TaS domains in the global feature space is described in Eq.~\ref{eq:TtoSglobal}. 
\begin{equation}
\normalsize
InfoNCE^{T, TaS}_{global} =  -log \frac{exp(sim(G^T_i, G^{TaS}_i)/\tau)}{\sum_{j=1}^{N}exp(sim(G^T_i, G^{TaS}_j)/\tau)} \\
-log \frac{exp(sim(G^{TaS}_i, G^T_i)/\tau)}{\sum_{j=1}^{N}exp(sim(G^{TaS}_i,G^T_j)/\tau)}, j \neq i
\label{eq:TtoSglobal}  
\end{equation}
The accumulated global domain adaptation loss in a mini-batch is now calculated in Eq.~\ref{eq:globalLoss}. 
\begin{equation}
    \normalsize
InfoNCE_{global}= InfoNCE^{S, SaT}_{global} \\
+ InfoNCE^{T, TaS}_{global}
\label{eq:globalLoss}
\end{equation}
Finally, the \emph{LGDA} model combines the local and global contrastive loss with the detection loss and the final loss function is now calculated as in Eq.~\ref{eq:totalLoss}: \begin{equation}
\normalsize
Total Loss= W_1 * InfoNCE_{global} + W_2 * InfoNCE_{local} \\
+ DWFL(x, p, y)
\label{eq:totalLoss}
\end{equation}

In the above Eq. \ref{eq:totalLoss}, the $W_1$ and $W_2$ denotes the weight we put on the two different modules.

\section{Experiments}
\label{sec:Exp}

We evaluated the proposed approach in the three largest annotated satellite image collections, DIOR \cite{2020LiObject}, DOTA2.0\cite{2021Ding}, and NWPU VHR-10 \cite{cheng2016survey} with four state-of-the-art domain adaptation models. DIOR is the data set for the source domain, and we used the ground-truth annotation for DIOR in our detection module and model evaluation. DOTA2.0 and NWPU VHR-10 are the target data set that adapts to the local and global domains, as presented in Table~\ref{tab:instance_dist}. We assume that the DOTA2.0 and NWPU VHR-10 annotations are unavailable at the object detection training time, and we use ground truth to evaluate the system's performance only at the testing time. In our experiments, we kept only common classes available in the DIOR and DOTA2.0 datasets. We assigned the same class label to each corresponding class in each dataset, as demonstrated for the reduced DIOR, DOTA2.0, and NWPU VHR-10 dataset in Table~\ref{tab:instance_dist}. We describe the experimental dataset in Subsection \ref{subsec:dataset}, the setup and implementation details in Section \ref{subsec:setup}, the performance comparison in Section \ref{subsec:measures}, and lastly, the ablation study in Section \ref{subsec:Ablation}.

\begin{figure}[!ht]
\centering
\includegraphics[width=0.50\textwidth]{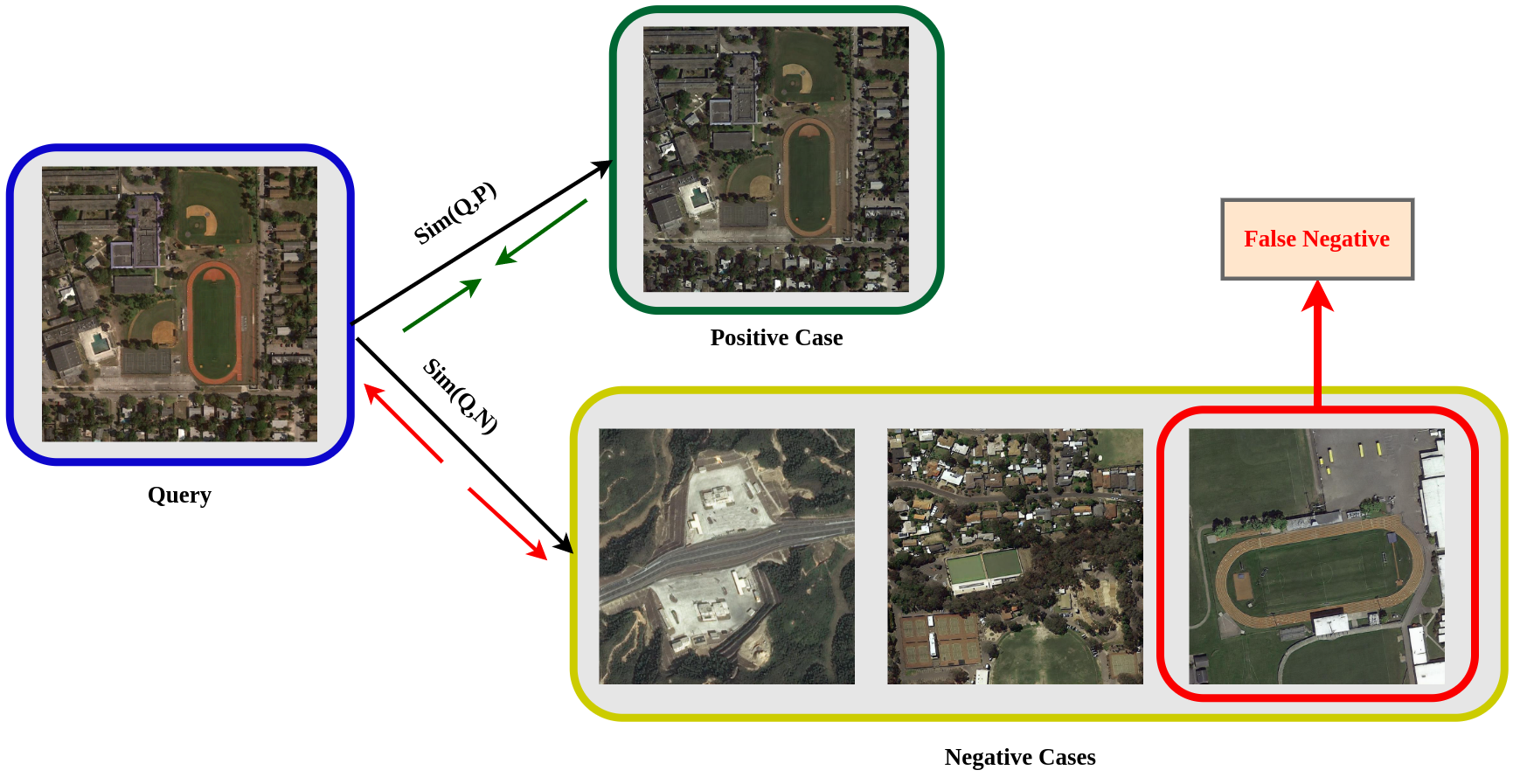}
    \caption{Example of False Negative occurrence in a highly imbalanced dataset.} \label{fig:bias}
\end{figure}

\begin{table}[htbp]
  \centering
  \normalsize
    \renewcommand{\arraystretch}{1.0}
  \begin{tabular}{ccccc}
    \toprule
    Class Name & \# of Ins. & \# of Ins. & \# of Ins.\\
    ~ & DIOR & DOTA & NWPU VHR-10\\
    \midrule
    Bridge & 176 & 1039 & 124\\
    Vehicle & 2079 & 85479 & 598\\
    Harbor & 254 & 5704 & 224\\
    Storage.T  & 2623 & 5416 & 655\\
    Stadium & 40 & 393 & -\\
    Baseball & 250 & 516 & 390\\
    Track & 138 & 417 & 163\\
    Basketball & 171 & 358 & 159\\
    Tennis  & 580 & 1662 & 524\\
    Airport & 25 & 153 & -\\
    \bottomrule
    \vspace{+0.5em}
  \end{tabular}
    \caption{Test set instance distribution statistics of the DIOR, DOTA2.0, and NWPU VHR-10 datasets across common categories.}
    \label{tab:instance_dist}
\end{table}

\subsection{Datasets}
\label{subsec:dataset}
The \textbf{DIOR} data set originally consisted of 24,500 Google Earth images from 80 countries. However, after selecting only common classes, the reduced dataset has 11,402 images. The images varied in quality and were captured in different seasons and weather conditions. The dataset boasts a wide range of spatial resolutions, object sizes, object orientations, and a diverse class distribution, as shown in Fig.~\ref{tab:instance_dist}. The spatial resolution of the images ranges from $[0.5m,30m]$, with each image measuring $800 \times 800$ pixels. The data set has 97,450 annotated objects, classified into ten classes \cite{2020LiObject}. Out of the 11,402 images, 10,888 are in the training set, and the remaining 512 images are in the testing set.

The \textbf{DOTA2.0} dataset comprises 2,430 overhead images sourced from Google Earth, Gaofen-2 (GF-2), and Jilin-1 (JL-1) satellites \cite{2021Ding}. The image sizes range from $800 \times 800$ to $29,200 \times 27,620$ pixels. The ground sample distance (GSD) in the data set ranges from 0.1 to 0.87 m, and each image contains an average of $~220$ objects. In the experiment, large images were split into overlapping tiles of $1024 \times 1024$ pixels with a 200-pixel overlap, resulting in 23,300 images from the original DOTA2.0 dataset. The reduced DOTA2.0 dataset has 11,551 images in the training set and 3,488 in the validation set, classified into ten classes (See Table \ref{tab:instance_dist}). In this paper, we interchangeably use DOTA and DOTA2.0 for referring to this dataset.

The \textbf{NWPU VHR-10} has in total 800 High-resolution images in the dataset, of which 715 images were collected from Google Earth, and the remaining are very-high-spatial-resolution pan-sharpened color infrared (CIR) images collected from the Vaihingen dataset \cite{cramer2010dgpf}. The GSD in the dataset ranges from 0.5 to 2m, and the image size range from $800 \time 560$ to $1267 \time 780$. Among the 800 High-resolution images, 650 are positive images with the available target in the annotation. In contrast, the rest of the images do not contain the target object considered a negative image. We only use positive images from the dataset for our Domain Adaptation task experiments. The reduced NWPU VHR-10 dataset has 450 images in the training and 200 images in the test set, distributed among eight classes (See Table \ref{tab:instance_dist}). In this paper, we interchangeably use NWPU VHR-10 and NWPU for referring to this dataset.

 \begin{table*}[!ht]
 \setlength\tabcolsep{2pt}
 \scriptsize
  \centering
    \renewcommand{\arraystretch}{1.2}
  \begin{tabular}{p{1.5cm}|p{2.0cm}|p{0.7cm}p{0.7cm}p{0.7cm}p{0.8cm}p{0.8cm}p{0.6cm}p{0.7cm}p{0.7cm}p{0.8cm}p{0.8cm}p{1.0cm}p{0.7cm}}\\ 
    Method & Detector+Backbone & Bridge & Vehicle & Harbor & Storage & Baseball & Track & B.Ball & Tennis & Stadium & Airport & DIOR $\rightarrow$ & DOTA\\
    ~ &  &  ~  & ~ & ~ & Tank & Field & Field & Court & Court & ~ & ~ & mAP & mAP\\ \hline
    Baseline \cite{zhou2021probabilistic} & CenNet2 ResNet50 & 10.5 & 8.9 & 42.1 & 40.6 & 46.5 & 31.4 & 46.7 & 74.2 & 0.0 & 28.3 & 64.8 & 32.1\\ 
    QueryDET \cite{yang2022querydet} & RetinaNet ResNet50 & 14.1 & 14.5 & 38.2 & 50.8 & 43.0 & 33.4 & 46.6 & 77.5 & 5.3 & 35.1 & 69.7 & 35.8\\
    EPM \cite{hsu2020every} & FCOS ResNet101 & 10.1 & 10.8 & 40.6 & 47.7 & 46.2 & 34.8 & 48.7 & 81.9 & 1.2 & 35.5 & 65.5 & 35.7\\
    MGADA \cite{zhou2022multi} & FCOS VGG16 & 13.1 & 10.8 & 45.9 & 48.5 & 46.0 & 37.7 & 50.1 & 84.3 & 0.0 & 37.2 & 66.9 & 37.3\\
    SAPNET \cite{li2020spatial} & FCOS ResNet50 & 10.9 & 11.0 & 23.5 & 24.4 & 35.3 & 27.8 & 32.2 & 74.1 & 0.0 & 22.7 & 54.7 & 26.1\\
    MGADA \cite{zhou2022multi} & F-RCNN ResNet101 & 15.9 & 14.0 & 48.1 & 46.5 & 47.6 & 39.3 & 52.6 & 87.2 & 1.8 & 37.9 & 72.6 & 39.4\\
    \textbf{SOD} & CenNet2 Darknet53 & 13.9 & 15.8 & 36.7 & 48.1 & 46.3 & 31.7 & 45.3 & 78.6 & 4.1 & 32.9 & 70.1 & 35.2\\
    \textbf{HeatDA} & CenNet2 Darknet53 & 14.4 & 17.4 & 39.1 & 50.9 & 46.1 & 35.5 & 48.0 & 80.1 & 4.8 & 37.6 & 70.1 & 37.4\\
    \textbf{LGDA*} & CenNet2 ResNet50 & 22.0 & 26.8 & 48.6 & 59.7 & 56.9 & 44.6 & 56.7 & 85.4 & 5.5 & 38.2 & \textbf{74.6} & \textbf{44.3} \\
    \textbf{LGDA} & CenNet2 Darknet53 & 24.5 & 27.9 & 51.3 & 62.0 & 59.2 & 47.9 & 58.6 & 87.8 & 6.1 & 38.7 & \textbf{76.9} & \textbf{46.7} \\ \hline
    Oracle & Baseline & 46.2 & 40.4 & 82.6 & 65.8 & 64.4 & 60.0 & 77.2 & 93.5 & 27.2 & 54.3 & \textbf{59.7} & \textbf{61.2}\\ 
  \end{tabular}
    \caption{Quantitative performance comparisons (mAP) across classes for DIOR → DOTA benchmark(IOU=0.5), where DIOR is considered as the source and DOTA as the target dataset. Class-wise performance is presented only for target dataset.}
    \label{tab:diortodota_classwise}
\end{table*}

\subsection{Experimental Setup and Implementation}
\label{subsec:setup}
 Source as Target (SaT) and Target as Source (TaS) synthesized data are created using the PyTorch implementation of the CycleGAN \cite{CycleGAN2017,isola2017image} network with the setup: the learning rate was 0.001; the number of training epochs was 2; load size was 800; and the crop size was 640. 

The proposed model for object detection architecture \emph{LGDA} is illustrated in Figure \ref{fig:LGDA} and is described in Section \ref{ssec:UDACons} is an extension of the CenterNet2 \cite{zhou2021probabilistic}  and SOD \cite{biswas2022small} model. The work in SOD uses Darknet53 as the backbone as it is shown to preserve better semantic information  from the small objects with the help of Cross-Stage-Partial(CSP) network than the residual-based feature extractor networks \cite{bochkovskiy2020yolov4,wang2020cspnet}; RPN heatmap-based approach to identify dense small objects and remove NMS; and the detection block is Faster-RCNN \cite{ren2015faster}.

We developed our code implementation by leveraging an open-source computer vision library \textbf{Detectron2} \cite{wu2019detectron2} and some part of \textbf{CenterNet2} \cite{zhou2021probabilistic}. We implemented two new DA modules for local and global domain adaptation using the contrastive learning technique. To train our \textbf{LGDA} model, we have resized all images to $800 \times 800$ pixels, and the mini-batch size in each epoch is set to 8. The loss of InfoNCE in Equation~\ref{eq:consloss} requires the same image from different domains as in the query and the positive case. To achieve this goal, we created a custom data loader in PyTorch that fetches the exact image of the different domains in a mini-batch. This custom data loader ensures that the Query and Positive examples are in the mini-batch sample during the training. In our experiments, eight is the size of the mini-batch, and the $8 \times 4 = 32$ images were fed into the \emph{LGDA} model in each mini-batch. The number of negative cases and the temperature were set to 7 and 0.12, respectively, for contrastive learning. Our research found that passing the S and SaT for the object detection task during the full training time makes the model over-fitted to the source domain. To eliminate this problem, we implemented random sample selection, randomly selecting 8 of 16 images from S and SaT and passing them on to the object detection module. Li et al. \cite{li2016attentive} found that Global Average Pooling (GAP) loses important context information when working with satellite images, so we replaced GAP with the bottleneck and MLP module for channel reduction. We have used NVIDIA 2 x RTX 6000 GPU with 49GB of memory, $11^{th}$ generation Intel® CoreTM i9-11900K @ 3.50GHz × 16 CPU, and 167GB of system memory to carry out all experiments.\\

\subsection{Performance Measures and Comparisons}
\label{subsec:measures}

We use Average Precision ($AP$), and Average Precision ($mAP$) to evaluate and compare the models. The precision ($P$), recall ($R$), F1 score ($F1$), and Mean Average Precision ($mAP$) are computed using Eq. \ref{eq:metric} and \ref{eq:AP}. True positives ($TP$) are instances correctly predicted by the model, false positives ($FP$) are instances missed by the model, and true negatives ($TN$) are instances incorrectly predicted by the model. We use an IOU threshold of 0.5 for all experiments, and the number of proposals per image is 256. Precision ($P$) represents the fraction of relevant instances recovered by the model, while recall ($R$) measures the fraction of relevant instances correctly identified by the model among all relevant instances. 

\begin{equation}
P=\frac{TP}{TP+FP},~R=\frac{TP}{TP+FN},~F1=\frac{2PR}{P+R}
\label{eq:metric}
\end{equation}

\begin{equation}
AP={\sum_{k=0}^{k=n-1} [R(k) - R(k+1)] * P(k)},\\
~mAP={\frac{1}{n}\sum_{k=1}^{k=n} AP(k)}
\label{eq:AP}
\end{equation}

The $F1$ score provides a single measure of the model's performance when given a class imbalance dataset. The mAP is calculated as shown in Eq.~\ref{eq:AP}, where $n$ is the number of classes in the test set, and $AP(k)$ is the Average Precision (AP) of class $k$ in the test set. Here, AP is the weighted sum of precision at each threshold ($n$ is the number of thresholds), and the weight is the increase in recall (Eq.~\ref{eq:AP}). 

 \begin{table*}[!ht]
 \setlength\tabcolsep{2pt}
  \centering
  \scriptsize
  \renewcommand{\arraystretch}{1.2}
  \begin{tabular}{p{1.7cm}|p{2.4cm}|p{0.9cm}p{0.9cm}p{0.9cm}p{1.0cm}p{1.0cm}p{0.8cm}p{0.9cm}p{0.9cm}p{1.2cm}p{0.8cm}}\\ 
    Method & Detector+Backbone & Bridge & Vehicle & Harbor & Storage & Baseball & Track & B.Ball & Tennis & DIOR $\rightarrow$ & NWPU\\
    ~ &  &  ~  & ~ & ~ & Tank & Field & Field & Court & Court & mAP & mAP\\ \hline
    Baseline \cite{zhou2021probabilistic} & CenNet2 ResNet50 & 32.5 & 27.1 & 76.5 & 48.4 & 30.3 & 66.1 & 63.6 & 67.0 & 66.8 & 52.0\\
    QueryDet \cite{yang2022querydet} & RetinaNet ResNet50 & 43.2 & 33.9 & 80.4 & 54.7 & 37.1 & 69.5 & 70.1 & 72.6 & 72.4 & 57.6\\
    EPM \cite{hsu2020every} & FCOS ResNet101 & 49.6 & 39.4 & 86.7 & 61.6 & 43.9 & 76.1 & 75.7 & 77.0 & 68.9 & 64.0\\
    MGADA \cite{zhou2022multi} & FCOS VGG16 & 48.8 & 37.1 & 85.5 & 59.6 & 42.3 & 76.8 & 72.6 & 77.2 & 66.5 & 62.6\\
    SAPNET \cite{li2020spatial} & FCOS ResNet50 & 35.8 & 22.6 & 70.1 & 41.4 & 25.5 & 60.1 & 56.5 & 61.4 & 60.8 & 46.7\\
    MGADA \cite{zhou2022multi} & F-RCNN ResNet101 & 54.2 & 43.9 & 88.5 & 65.0 & 49.9 & 77.2 & 76.6 & 77.0 & 69.2 & 66.8\\
    \textbf{SOD} & CenNet2 Darknet53 & 47.5 & 36.4 & 83.9 & 54.5 & 38.5 & 69.0 & 72.6 & 72.8 & 71.5 & 59.7\\
    \textbf{HeatDA} & CenNet2 Darknet53 & 48.8 & 38.2 & 85.5 & 56.9 & 49.5 & 70.6 & 74.5 & 73.0 & 71.9 & 62.1\\
    \textbf{LGDA*} & CenNet2 ResNet50 & 54.0 & 43.9 & 90.5 & 62.0 & 53.9 & 76.2 & 78.6 & 81.0 & \textbf{73.8} & \textbf{67.9} \\
    \textbf{LGDA} & CenNet2 Darknet53 & 57.6 & 51.5 & 90.4 & 65.7 & 60.1 & 79.8 & 81.3 & 84.7 & \textbf{74.6} & \textbf{71.4} \\ \hline
    Oracle & Baseline & 69.5 & 60.9 & 96.5 & 79.4 & 64.3 & 96.1 & 93.6 & 97.8 & \textbf{60.6} & \textbf{84.4}\\ 
  \end{tabular}
    \caption{Quantitative performance comparisons (mAP) across classes for DIOR → NWPU VHR-10 benchmark(IOU=0.5), where DIOR is considered as the source and NWPU as the target dataset. Class-wise performance is presented only for the target dataset.}
    \label{tab:diortonwpu_classwise}
\end{table*}

\textbf{Object Detection Comparisons:} To compare our proposed model with recent state-of-the-art (SOTA) models, we set a lower-bound and an upper-bound on the performance for each dataset. We use \emph{CenterNet2} as the baseline/ lower-bound for comparisons, where we use annotations from only the \emph{source} datasets during the training phase to evaluate the target dataset. On the other hand, \emph{oracle} is the upper bound, which uses \emph{CenterNet} as the detection model and uses annotation from \emph{Target} dataset while training to evaluate the target dataset. We choose CenterNet2 as our baseline model because of its ability to work better with small and dense objects leveraging the power of heatmap-based RPN. We use our extended version of the SOD pipeline for small object detection architecture.  We have compared our \emph{SOD} model performance with a recent small object detection SOTA model \emph{QueryDet} \cite{yang2022querydet} in Table \ref{tab:diortodota_classwise} and \ref{tab:diortonwpu_classwise}. We can see that our model performs very close to the \emph{QueryDet} method for the DOTA dataset, and outperforms the \emph{QueryDet} by 2.1\% of mAP on the NWPU VHR-10 dataset. Moreover, using 12GB GPU memory \emph{QueryDet} trains 2 images per batch whereas using \emph{SOD} we can set batch size up to 8 images, so we decided to continue further experiments using SOD as the small object pipeline. As the domain adaptation SOTA models, we used feature alignment DA methods such as MGADA \cite{zhou2022multi} and a spatial attention-based domain adaptation network SAPNet \cite{li2020spatial} for the performance measurements. Also, we introduced EPM \cite{hsu2020every}, a domain adaptation framework that accounts for each pixel via predicting pixel-wise centeredness and objectness for state-of-the-art comparisons. Later we keep track of the performance improvement of our new proposed model \emph{LGDA} and try to minimize the gap between \emph{LGDA} and \emph{oracle} for domain adaptation task. Except for the \emph{oracle}, all other comparing models use annotation from only the \emph{source} dataset and images from the \emph{source} and \emph{target} datasets during the training phase. Next, we evaluate the performance of the models based on the test set of both \emph{source} and \emph{target} datasets. By this, we show that our proposed UDA models perform satisfactorily on the target domain, and there is no performance degradation in the source domain due to induced noise from the DA operation. 

We start our DA evaluation in Table \ref{tab:diortodota_classwise}, using DIOR as the source and DOTA as the target. Table \ref{tab:diortodota_classwise} shows that the CenterNet2-based baseline model gives an mAP of 64.8\% on the source and 32.1\% mAP on the target dataset. We improve the baseline model with the integration of \emph{Custom Focal Loss}, \emph{Difficulty Estimation Block} and \emph{Strong Backbone} as illustrated in Figure \ref{fig:SOD} and propose a new model \emph{SOD} verified to work better on satellite imagery \cite{biswas2022small}. Table \ref{tab:diortodota_classwise} shows significant improvements on both datasets for small objects such as harbors, vehicles, and storage tanks. The range of $\alpha'_c$ values in the DIOR dataset is 0.2 to 0.79, and the range of $\alpha'_c$ values in the DOTA2.0 dataset is 0.15 to 0.96, which represents a very tight scaling factor for \emph{FL} in both data sets. The SOD model also shows reasonable promise in performance improvement with a gain of 6.0\% and 3.0\% of mAP for the DIOR and DOTA datasets, respectively, by dealing with small objects and challenging images. The first step towards DA operation was introducing transfer learning and using CycleGAN-generated composite target images for the DA training. We name the model as \emph{HeatDA}, which uses the target domain pixel-level context and gives 5.3\% mAP improvement on the target dataset. Our final proposed model \emph{LGDA} is an extension of \emph{HeatDA} model by adding local and global domain adaptation modules. Compared to the baseline model, our \emph{LGDA} model gains 12.2\% and 14.6\% of mAP on the target dataset using ReseNet50 and CSP-Darknet53 backbone, respectively. The detection results from \emph{LGDA} model are illustrated in Figure \ref{fig:detections}, where the proposals generated from heatmaps are shown in Figure \ref{fig:detections}(a) and (c), and the object detection performance is shown in Figure \ref{fig:detections}(b) and (d). Figure \ref{fig:detections} also shows us the value of image difficulty for a particular image. It is evident that an image with a higher number of objects and pixel diversity score more difficulty value than an image with fewer objects and less pixel variations. Among the other SOTA models, MGADA performs best with an mAP of 39.4\%, and SAPNet gives the lowest mAP of 26.1\%. However, Our proposed novel contrastive learning method with a small object-focused pipeline helps us to outperform other SOTA models by a minimum margin of 7.3 \% on the target dataset.

\begin{table}[!ht]
 \setlength\tabcolsep{1pt}
  \centering
  \normalsize
  \renewcommand{\arraystretch}{1.2}
  \begin{tabular}{l|c|cccc|cc}
  \toprule
    Method & Backbone & TL & LDA & GDA & DWFL & DOTA & NWPU\\ \hline
    Baseline &  &  &  &  &  & 32.1 & 52.0\\
    w/DWFL &  &  &  &  & \checkmark & 35.2 & 59.7\\
    w/TL &  & \checkmark &  &  & \checkmark & 37.4 & 61.8\\
    w/LDA & DarkNet53 &  & \checkmark &  & \checkmark & 40.6 & 64.2\\
    w/GDA &  & & &\checkmark & \checkmark & 41.5 & 66.6\\
    LGDA &  & \checkmark &\checkmark & \checkmark & \checkmark & 46.7 & 71.4\\
    \bottomrule
  \end{tabular}
  \vspace{+0.5em}
    \caption{Ablation study for our proposed LGDA method. Here, TL= Transfer Learning, LDA= pixel-level local domain adaptation, and GDA= object-level global domain adaptation.}
    \label{tab:ablation_LGDA}
\end{table}

\begin{table*}[htbp]
 \setlength\tabcolsep{1pt}
  \centering
  \scriptsize
  \renewcommand{\arraystretch}{1.2}
\begin{tabular}{c|cccc|cccc|cccc}
  & \multicolumn{4}{c|}{Precision} & \multicolumn{4}{|c|}{Recall} & \multicolumn{4}{|c}{F1} \\ \hline 
Dataset & CenterNet2 & SOD & HeatDA & LGDA & CenterNet2 & SOD & HeatDA & LGDA & CenterNet2 & SOD & HeatDA & LGDA \\ \hline \hline 
DOTA & 32.1 & 35.2 & 37.4 & \textbf{46.7} & 46.9 & 47.7 & 50.4 & \textbf{60.2} & 38.1 & 46.1 & 42.9 & \textbf{52.6}\\
NWPU VHR-10 & 52.0 & 59.4 & 61.8 & \textbf{71.4} & 58.8 & 63.5 & \textbf{67.9} & 66.8 & 61.1 & 61.3 & 64.7 & \textbf{69.0} \\
\end{tabular}
 \caption{Comparison of Precision, Recall, and F1 score between our proposed models for the DOTA and NWPU VHR-10 target datasets.}
\label{tab:PR_comp_dior2dota}
\end{table*}

\begin{figure*}
    \centering
    \includegraphics[width=0.95\textwidth]{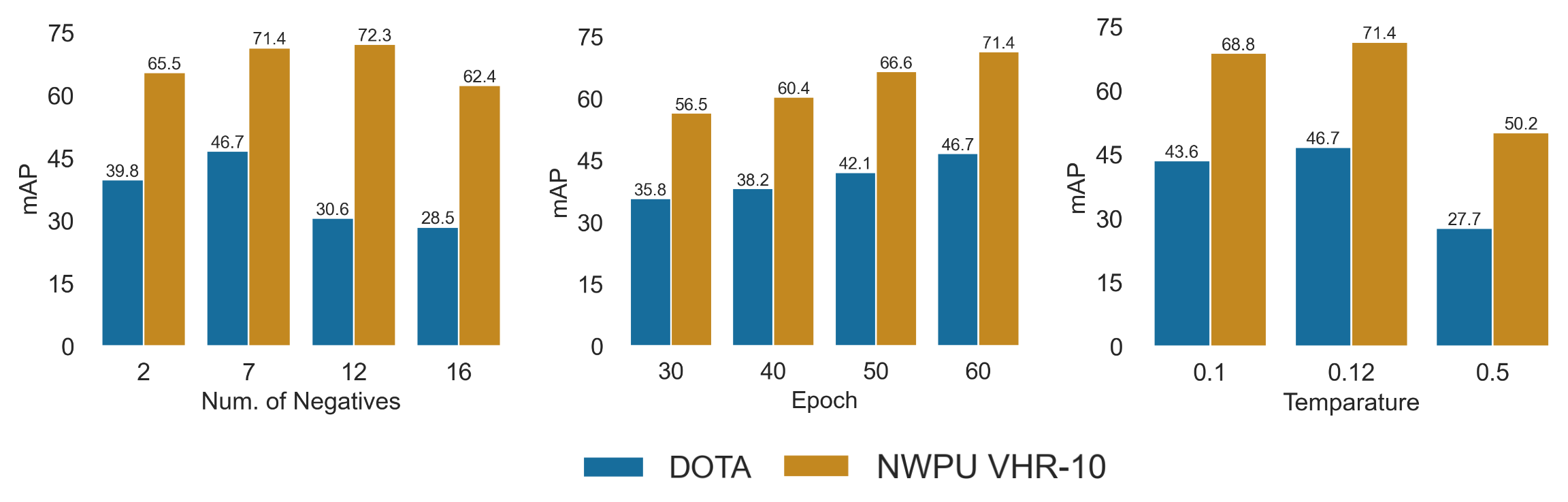}
    \caption{Ablation Study on Hyper-parameters: (a) mAP vs. Negative Examples, (b) mAP vs. Epochs, and (c)  mAP vs.Temperature.}
    \label{fig:ablation_hparam}
\end{figure*}

Next, we use Table \ref{tab:diortonwpu_classwise} to demonstrate the DA performance and SOTA comparison for DIOR and NWPU VHR-10 datasets. The DIOR and NWPU VHR-10 are two high-variability image datasets, as shown in Table \ref{tab:instance_dist} captured from satellites. Here, we evaluate target dataset performance over eight different categories. It is observed during experiments that we have not only demonstrated outstanding performance on the target dataset but also we have attained a notable 74.6\% mAP(refer to Table \ref{tab:diortonwpu_classwise}) on the source dataset. Our baseline CenterNet2 method trained on only source dataset archives 52.0\% of mAP, whereas our LGDA method achieves 71.4\% of mAP using contrastive learning with local and global domain adaptation. Also, we have a +4.6\% gain margin compared to the best state-of-the-art \emph{MGADA} method.
Moreover, compared to the baseline model, we were able to shrink the performance gap between the oracle and our model from 32.4\% to 13.0\% using the Local-Golbal DA. Table \ref{tab:diortonwpu_classwise} and Figure \ref{fig:detections}(b) and (d) demonstrate the effectiveness of our method in detecting objects from challenging and less frequent categories, including track, bridge, and basketball fields. Table \ref{tab:diortonwpu_classwise} further illustrates that a meticulously designed backbone can augment the performance by approximately +3.5\% on the target domain, mainly when dealing with densely populated objects. 

Finally, we perform an in-depth performance analysis in Table \ref{tab:PR_comp_dior2dota} using Precision, Recall, and F1-Score for the four proposed models. For the DOTA dataset, we gain the optimal result in all metrics using the final version of the \emph{LGDA} model, with a Precision, Recall, and F1-Score of 46.7, 60.2, and 52.6, respectively. However, for NWPU, we got the optimal value of Precision and F1-Score from the \emph{LGDA} model and Recall from the \emph{HeatDA} model.
 
\subsection{Ablation study}
\label{subsec:Ablation}

In this section, we first perform an ablation study on each component of our proposed \emph{LGDA} method to demonstrate the effectiveness of each element, as shown in Table \ref{tab:ablation_LGDA}. Our ablation study is carried out on two target datasets: DOTA and NWPU VHR-10. Next, we perform an ablation study on each important hyper-parameters; the summary of the ablation study is illustrated in Figure \ref{fig:ablation_hparam}. Table \ref{tab:ablation_LGDA} shows that integrating Difficulty Weighted Focal Loss (DWFL) was crucial for our proposed model as we made 3.1\% and 7.7\% increase in mAP for DOTA and NWPU, respectively. The intermediate version of our proposed model is \emph{HeatDA}, where we investigate the amount of efficiency we can leverage from Transfer Learning, Synthetic images, and DWFL. The Synthetic image generated from GAN networks provides primary pixel-level color and texture information, and we can notice a slight gain of mAP in Table \ref{tab:ablation_LGDA} for both target datasets. Then we integrate contrastive learning into our DA process and propose two new modules; the first is for aligning local features between the source and target domain, and the latter is for aligning a more abstract view of features with a high-receptive field. The local feature alignment with contrastive learning (LDA) helped us gain over 8\% and 12\% mAP on DOTA and NWPU VHR-10 by putting a weight of 0.1 on the loss function. On the other side, from the GDA module, we even get better results than the LDA with a weight of 0.01 on contrastive learning. Finally, integrating all small modules, we propose the LGDA method, which achieves 46.7\% and 71.4\% of mAP on DOTA and NWPU VHR-10 target datasets, respectively.

Next, we look for the optimal value for the number of negative examples for the hyper-parameters study, as shown in Figure \ref{fig:ablation_hparam}(a). We started our experiments with a value of 2, and we can see that the model did not perform well with fewer negative examples in the contrastive loss. We increased the value to 7 to generalize learning and recorded our best performance in the DOTA target dataset. Increasing the negative examples further did not help us learn in target datasets due to the imbalanced nature of the data set. The vehicle class dominates our DOTA data set, as shown in Table \ref{tab:instance_dist}. Increasing the number of negative examples also increases the chances of getting false negative (FN) examples, as shown in Figure \ref{fig:bias} in contrastive learning. In highly imbalanced datasets such as DOTA2.0, the FN example makes the model biased toward a particular class, and the model's overall performance degrades significantly. However, there was a slight improvement for the NWPU VHR-10 dataset with negative example 12, as NWPU is much more balanced than DOTA. The adverse effect of bias is evident in Figure \ref{fig:ablation_hparam}(a) when trained with 16 negative examples. After careful inspection and to reduce computational expense, we set the number of negative measures equal to 7 for both datasets.

Second, our baseline model was trained for 30 epochs, and the LGDA model was trained for 30 more epochs as we added two more loss functions for contrastive learning. As illustrated in Figure \ref{fig:ablation_hparam}(b), our experiments found that training for more epochs does not significantly improve performance. Therefore, all results were recorded with 60 training epochs. Lastly, the temperature value in contrastive loss is susceptible and small changes in value can drastically change the outcome. This is evident in Figure \ref{fig:ablation_hparam}(c); placing a 50\% penalty on contrastive loss dramatically reduces performance, and using a 12\% penalty shows the optimal result on target datasets.

\section{Conclusion and Future Work}
\label{sec:Con}
Object detection in aerial images is one of the most challenging tasks in computer vision research because many small and overlapped objects exist in the photos. The success of DNN object localization depends on a large amount of annotated training data and a reliable feature extractor module in the pipeline. This paper presents a robust feature extractor that captures balanced low- and high-level features for small objects. Next, we offer the heat-map-based region proposal module to grab small things better. The domain gap in satellite images is more significant than in consumer images because of weather conditions, geographic changes, and camera orientations. We perform progressive domain alignment by creating two intermediate domains, w.r.t. source and target datasets. The proposed method \emph{LGDA} performed exceptionally well with more than 60\% mAP for several classes such as \emph{storage tank}, \emph{harbor}, and \emph{tennis court} in the DOTA and NWPU VHR-10 target data sets. We also use contrastive learning to adapt to local and global domains. Careful selection of the training pipeline, the number of negative samples, the down-sampling strategy, and the temperature value can improve the effectiveness of contrastive learning. Finally, we validate our approach in two challenging high-variability target datasets that showed significant performance gain over available state-of-the-art methods. For the DOTA and NWPU VHR-10 target datasets, we outperformed the latest state-of-the-art \emph{MGADA} method by +7.3\% and +4.6\% mAP, respectively. Next, we plan clustering-based pseudo-labeled for target objects, de-biased instance-level domain adaptation, and unknown class discovery for satellite images.

\bibliographystyle{unsrt}  

\end{document}